\def\year{2019}\relax
\documentclass[letterpaper]{article} %DO NOT CHANGE THIS
\usepackage{aaai19}  %Required
\usepackage{times}  %Required
\usepackage{helvet}  %Required
\usepackage{courier}  %Required
\usepackage{url}  %Required
\usepackage{graphicx}  %Required
\usepackage{amssymb}
\usepackage{bm}
\usepackage{tabu}
\usepackage{booktabs}
\usepackage{multirow}
\usepackage{comment}
\begin{document}
\title{A Deep Sequential Model for Discourse Parsing on Multi-Party Dialogues}
\author{
Zhouxing Shi \and Minlie Huang\thanks{Corresponding author: Minlie Huang.}\\
Dept. of Computer Science \& Technology, Tsinghua University, Beijing 100084, China\\
Institute for Artificial Intelligence, Tsinghua University (THUAI), China\\
Beijing National Research Center for Information Science and Technology, China\\
zhouxingshichn@gmail.com,
aihuang@tsinghua.edu.cn
}
\maketitle
\begin{abstract}
Discourse structures are beneficial for various NLP tasks such as dialogue understanding, question answering, sentiment analysis, and so on. This paper presents a deep sequential model for parsing discourse dependency structures of multi-party dialogues. The proposed model aims to construct a discourse dependency tree by predicting dependency relations and constructing the discourse structure jointly and alternately. It makes a sequential scan of the \emph{Elementary Discourse Units (EDUs)}\footnote{A discourse can be segmented into clause-level units called \emph{Elementary Discourse Units (EDUs)} which are the most fundamental discourse units in discourse parsing. Following previous work such as \cite{li2014text,li2014recursive}, we also assume that EDU segmentations are preprocessed.}
in a dialogue. For each EDU, the model decides to which previous EDU the current one should link and what the corresponding relation type is. The predicted link and relation type are then used to build the discourse structure incrementally with a structured encoder. During link prediction and relation classification, the model utilizes not only \textit{local} information that represents the concerned EDUs, but also \textit{global} information that encodes the EDU sequence and the discourse structure that is already built at the current step. Experiments show that the proposed model outperforms all the state-of-the-art baselines.

\end{abstract}

\section{Introduction}

Discourse parsing is to identify relations between discourse units and to discover the discourse structure that the units form \cite{li2016discourse}. 
Previous studies have shown that discourse structures are beneficial for various NLP tasks, including 
dialogue understanding \cite{STACcorpus,Takanobu2018Weakly}, question answering \cite{verberne2007evaluating}, information retrieval \cite{seo2009online}, and sentiment analysis \cite{cambria2013new,bhatia2015better}.

\begin{figure}[ht]
	\centering
	\includegraphics[width=8.0cm]{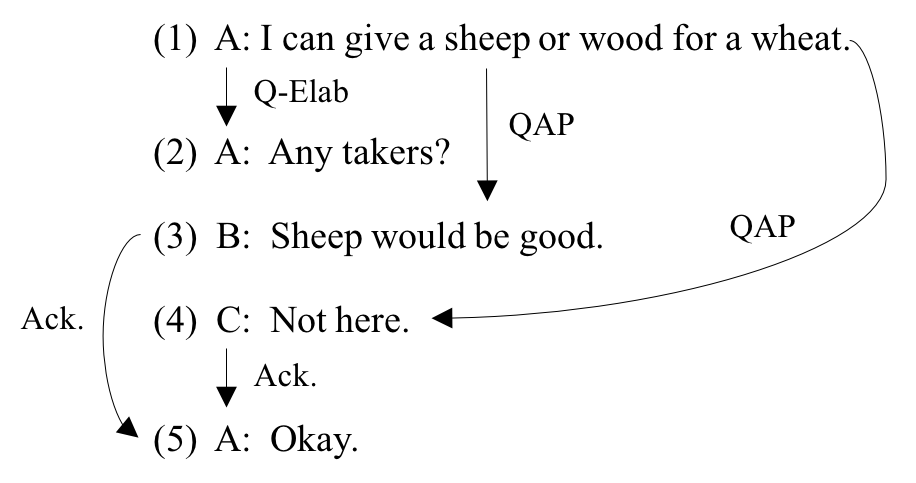}
    \caption{A multi-party dialogue example with its discourse structure from the STAC Corpus \cite{STACcorpus}, where ``Q-Elab'' is short for ``Question-Elaboration'', ``QAP'' for ``Question-Answer-Pair'', and ``Ack.'' for ``Acknowledgement''.}
    \label{dialogue_example}
\end{figure}

Many approaches have been proposed for discourse parsing based on Rhetorical Structure Theory (RST) \cite{mann1988rhetorical}. 
However, RST is designed for written text and only allows discourse relations to appear between adjacent discourse units, and thus is inapplicable for multi-party dialogues \cite{afantenos2015discourse} since multi-party dialogue data have more complex discourse structures in nature.
 RST is \emph{constituency-based}, where related adjacent discourse units are merged to form larger units recursively, resulting in a hierarchical tree structure \cite{li2014recursive}. By contrast,  \emph{dependency-based} structures, where EDUs are directly linked without forming upper-level structures, are more applicable for multi-party dialogues. It is because multi-party dialogues have immediate relations between non-adjacent discourse units and the discourse structures are generally non-projective\footnote{A discourse structure is \emph{non-projective} if it is impossible to draw the relations on the same side without crossing \cite{mcdonald2005non}. As the non-projective example shown in Figure \ref{dialogue_example}, $1\rightarrow 4$ and $3\rightarrow 5$ have to be drawn on two sides to avoid crossing.} 
 \cite{morey2018dependency}.
Therefore, the focus of this paper is on parsing \emph{dependency structures} of multi-party dialogues.
Figure \ref{dialogue_example} shows an example of a multi-party dialogue and its dependency structure, where three speakers (A, B, C) are conversing in an online game. 

Prior state-of-the-art approaches for discourse dependency parsing commonly adopt a pipeline framework: first estimating the local probability of the dependency relation between each combination of two EDUs, and then constructing a discourse structure with decoding algorithms such as maximum spanning tree or integer linear programming \cite{muller2012constrained,li2014text,afantenos2015discourse,perret2016integer}, based on the estimated probabilities.
However, these approaches have two drawbacks. 
\textbf{First}, the probability estimation of each dependency relation
between two EDUs only relies on the local information of these two considered EDUs. 
\textbf{Second}, dependency prediction and discourse structure construction are separated in two stages, thereby dependency prediction cannot utilize the information from the predicted discourse structure for better dependency parsing, and in return, worse dependency prediction degrades the construction of the discourse structure.

To address these drawbacks, we propose a deep sequential model for discourse parsing on multi-party dialogues. 
This model constructs a discourse structure incrementally by predicting dependency relations and building the structure jointly and alternately.
It makes a sequential scan of the EDUs in a dialogue. 
For each EDU, the model decides to which previous EDU the current one should link and what the relation type is. 
Such dependency prediction relies on not only  local information that encodes the two concerned EDUs, but also global information that encodes the EDU sequence and the discourse structure that is already built at the current step. 
The predicted link and relation type, in return, are used to build the structure incrementally with a structured encoder. In this manner, the model predicts dependency relations and constructs the discourse structure jointly and alternately.   

In summary, we make the following contributions:
\begin{itemize}
    \item We propose a deep sequential model for discourse parsing on multi-party dialogues. The model predicts dependency relations and constructs a discourse structure jointly and alternately.
    
    \item We devise a prediction module that fully utilizes local information that encodes the concerned units, and also global information that encodes the EDU sequence and the currently constructed structure. 
   
    \item We devise a structured encoder for representing structured global information, and propose a \emph{speaker highlighting mechanism} to utilize speaker information and enhance dialogue understanding.
\end{itemize}

\section{Related Work}

Most previous work for discourse parsing is based on Penn Discourse TreeBank (PDTB) \cite{prasad2007penn} or Rhetorical Structure Theory Discourse
TreeBank (RST-DT) \cite{mann1988rhetorical}. 
PDTB focuses on shallow discourse relations but ignores the overall discourse structure \cite{yang2018scidtb}, while in this paper we aim to parse discourse structures.
As for RST, there have been many approaches including transition-based methods  \cite{braud2017cross,wang2017two,yu2018transition} and those involving CYK-like algorithms \cite{joty2015codra,li2016discourse,liu2017learning} or greedy bottom-up algorithms \cite{feng2014linear}. 
However, constituency-based RST does not allow non-adjacent relations, which makes it inapplicable for multi-party dialogues.
By contrast, in this paper, we aim to parse non-projective dependency structures, where dependency relations can appear between non-adjacent EDUs.

There have been some approaches proposed for parsing discourse dependency structures in two stages.
These approaches first predict the local probability of dependency relation for each possible combination of EDU pairs, and then apply a decoding algorithm to construct the final structure. \cite{muller2012constrained,li2014text,afantenos2015discourse} used Maximum Spanning Trees (MST) to construct a dependency tree, and \cite{muller2012constrained} also attempted $A^*$ algorithm but did not achieve better performance than MST.
\cite{perret2016integer} further used Integer Linear Programming (ILP) to construct a dependency graph. 
However, these approaches predict the probability of a dependency relation only with the local information of the two considered EDUs, while the constructed structure is not involved.
By contrast, our sequential model predicts dependency relations and constructs the discourse structure jointly and alternately, and utilizes the currently constructed structure in dependency prediction.

Although transition-based approaches for discourse dependency parsing have been proposed by \cite{jia2018improved,jia2018modeling} which also construct dependency structures incrementally, they still underperform the approach using MST by \cite{li2014text}. It is because these transition-based local approaches do not investigate other possible links when predicting a dependency relation as argued by \cite{jia2018modeling}, and they are limited to predict projective structures.
Therefore, these approaches are inapplicable for multi-party dialogues.
By contrast, our sequential model predicts the parent of each EDU in the dependency tree by comparing all its preceding EDUs, and it can predict non-projective structures which are necessary for multi-party dialogues.

Moreover, state-of-the-art approaches for discourse dependency parsing as mentioned above still rely on hand-crafted features or external parsers. 
Neural networks have recently been widely applied in various NLP tasks, including RST discourse parsing \cite{li2016discourse,braud2017cross} and dialogue act recognition \cite{kumar2018dialogue,chen2018dialogue}. 
And \cite{jia2018improved,jia2018modeling} also applied neural networks in their transition-based dependency parsing models. 
In this paper, we adopt hierarchical Gated Recurrent Unit (GRU) \cite{cholearning} encoders to compute discourse representations.

\section{Methodology}

\begin{figure*}[ht]
	\centering
	\includegraphics[width=17.5cm]{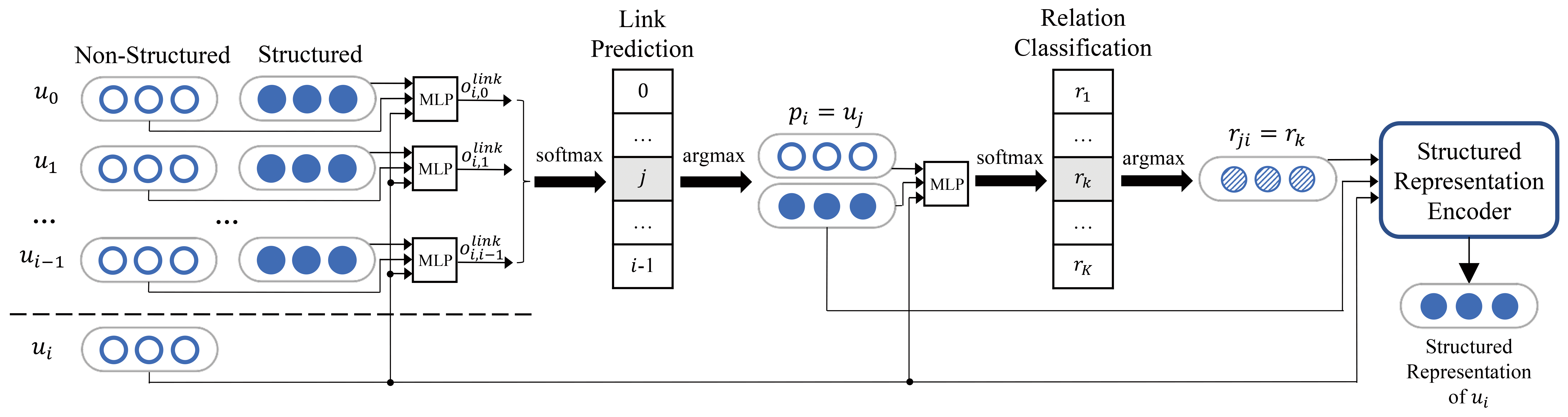}
	\caption{Illustration of the model which consists of modules for link prediction, relation classification, and structured representation encoding. For the current EDU $u_i$, link prediction estimates a distribution over its preceding EDUs, relation classification estimates a distribution over relation types, and the structured encoder updates the structured representation of $u_i$ using representations of $u_i$ and $p_i$ and the embedding of the predicted relation type $r_{ji}$. 
	Non-structured representation encoding is performed before the prediction process and is omitted from the illustration.
	}
	\label{model_overview}
\end{figure*}

\subsection{Problem Definition}
We formulate the discourse dependency parsing problem on a  multi-party dialogue as follows:
given a dialogue that has been segmented into a sequence of EDUs $u_1,u_2,\cdots,u_n$
, together with an additional dummy root $u_0$\footnote{The dummy root is for the convenience of problem definition following \cite{li2014text}.
},
the goal is to predict dependency links and the corresponding relation types $\{ (u_j,u_i,r_{ji})| j\neq i \} $
between the EDUs, where $(u_j,u_i,r_{ji})$ stands for a link of relation type $r_{ji}$ from $u_j$ to $u_i$.
The predicted dependency relations should constitute a Directed Acyclic Graph (DAG) and there should be no relation linked to $u_0$.

The discourse structure predicted by our model is a dependency tree, which is a special type of DAG\footnote{We found that the proportion of EDUs with multiple incoming relations is quite limited (less than 6.4\%) in the dataset we used. Nevertheless, our model can be easily extended to predict a more general DAG when necessary.}.
The model makes a sequential scan of the EDUs $u_1,u_2,\cdots,u_n$. For the current EDU $u_i$, the model predicts a dependency link by estimating a probability distribution as follows:
\begin{equation}
\mathcal{P}(u_j|u_i,\mathcal{T}_i, 0 \leq j \leq i-1)
\end{equation}
where $\mathcal{T}_i=\{(u_l,u_k,r_{lk})|  0 \leq l < k \leq i-1 \}$ is the set of dependency relations that are already predicted before the current step $i$.  This is so-called {\it link prediction} in our model.
Similarly, the model predicts the relation type for a predicted link $u_j \rightarrow u_i (j<i)$ with the following distribution:
\begin{equation}
\mathcal{P}(r_{ji}|u_j \rightarrow u_i,\mathcal{T}_i)
\end{equation}
where $r_{ji}\in\{r_1,r_2, \cdots, r_K\}$, $r_{k}(1\leq k\leq K)$ is a relation type and $K$ is the number of relation types. This is so-called {\it relation classification}.

\subsection{Model Overview}

Our model first computes the non-structured representations of the EDUs with hierarchical Gated Recurrent Unit (GRU) \cite{cholearning} encoders.
These non-structured representations are used for predicting dependency relations and encoding structured representations.
Next, the model makes a sequential scan of the EDUs and has the following three steps as illustrated in Figure \ref{model_overview} when it handles EDU $u_i$:
\begin{enumerate}
    \item \textbf{Link prediction}: predict the parent node $p_i$ of EDU $u_i$ with a link predictor which utilizes not only non-structured representations, but also structured representations that encode the predicted structure before $u_i$. 
    Specifically, we compute a score between the current EDU $u_i$ and each linking candidate $u_j(j<i)$ with an MLP. These scores are then normalized to a distribution over the previous EDUs $\{u_0,u_1,\cdots,u_{i-1}\}$ with $softmax$, from which we can take the linked EDU with the largest probability.
    
    \item \textbf{Relation classification}: predict the relation type between $p_i$ (assume $p_i=u_j$) and $u_i$ with a relation classifier. Similar to link prediction, the relation classifier leverages both non-structured and structured representations. Discourse representations of $u_j$ and $u_i$ are fed into an MLP to obtain a distribution over relation types.
    The relation type $r_{ji}$ is taken with the largest probability.
    
    \item \textbf{Structured representation encoding}: compute the structured representation of $u_i$ with a structured representation encoder which encodes the predicted discourse structure. Specifically, the relation embedding of $r_{ji}$, the non-structured representation of $u_i$, and the structured representation of $p_i=u_j$, are fed into the encoder to derive the structured representation of $u_i$.    
    
\end{enumerate}

Afterwards, the model moves to the next EDU $u_{i+1}$ and performs the above three steps iteratively until the end of the dialogue. In this manner, dependency prediction and discourse structure construction are performed jointly and alternately, and the discourse structure is built incrementally.

\subsection{Discourse Representations}

In our model, we use two categories of discourse representations: local representations and global representations.
Local representations are non-structured and encode the local information of EDUs individually.
And global representations encode the global information of the EDU sequence or the predicted discourse structure.
These representations are taken as the input for link prediction and relation classification. 
In return, the predicted links and relation types are used to build structured global representations incrementally.

\subsubsection{Local Representations}

For each EDU $u_i$, a bidirectional GRU (bi-GRU) encoder is applied on the word sequence, and the last hidden states in two directions are concatenated as the local representation of $u_i$, denoted as $\bm{h}_i$.

\subsubsection{Non-structured Global Representations}

Non-structured global representations encode the EDU sequence in a dialogue.
The local representations of the EDUs $\bm{h}_0,\bm{h}_1,\cdots,\bm{h}_n$ are taken as input to a GRU encoder and the hidden states are viewed as the \emph{non-structured global representations} of the EDUs, denoted as $\bm{g}^{NS}_0,\bm{g}^{NS}_1,\cdots,\bm{g}^{NS}_n$.

\begin{figure}[ht]
	\centering
	\includegraphics[width=8.0cm]{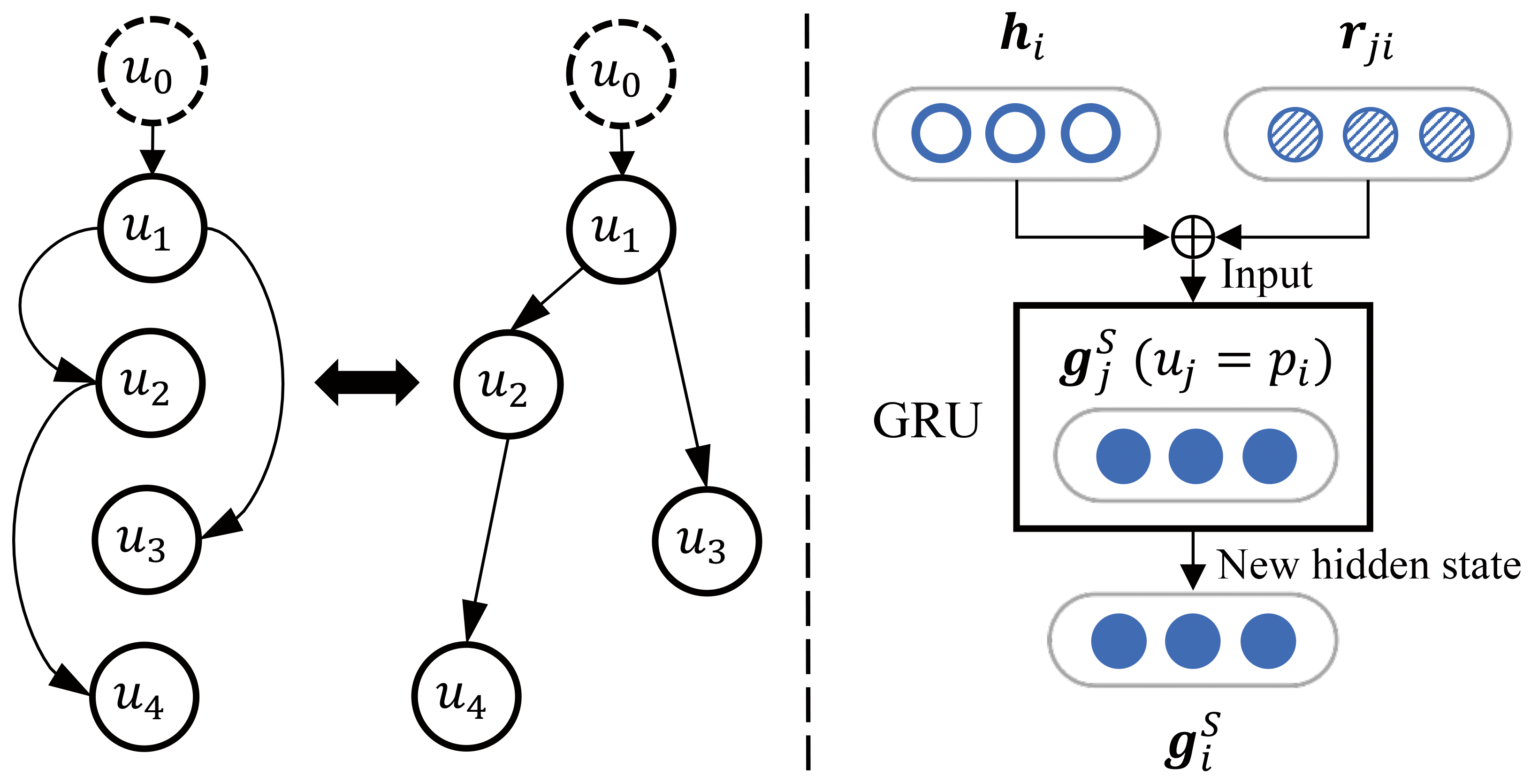}
	\caption{An example dependency tree (left) and the structured encoder (right), where $\bm{h}_i$ is the local representation of EDU $u_i$, $\bm{g}_i^S$ and $\bm{g}_j^S$ are structured representations, $\bm{r}_{ji}$ is the relation embedding, and $u_j=p_i$ is the parent of $u_i$.}
	\label{structured}
\end{figure}

\subsubsection{Structured Global Representations}

\noindent The structured representations encode dependency links and the corresponding relation types to fully utilize the global information of the predicted structure.
Note that there is exactly one path from the root to each EDU on the predicted dependency tree, and the path represents the development of the dialogue.
We apply a structured encoder to these paths to obtain \emph{structured global representations} (or \emph{structured representations} briefly) of the EDUs.
For instance, to obtain the structured representation of $u_4$ as shown in Figure \ref{structured}, the structured encoder is applied to the path $u_0\rightarrow u_1\rightarrow u_2 \rightarrow u_4$, and the hidden state at $u_4$ is treated as its structured representation.
The structured representations are computed incrementally. We compute the structured representation of $u_i$ once its parent and the corresponding relation type are decided.

In addition, when predicting a dependency relation linking from $u_j$ to $u_i$, it is beneficial to highlight previous utterances from the same speaker as that of $u_i$. Because it helps the model to better understand the development of the dialogue involving this speaker, which can improve the prediction of the dependency related to $u_i$. 
For instance, if we consider a dependency path like $\cdots \rightarrow u_k(a)\rightarrow \cdots \rightarrow u_j(b) \rightarrow u_i(a)$ where $a,b$ are speaker identifiers, when predicting the dependency link between $u_j$ and $u_i$, it is beneficial to highlight the previous dialogue history $u_k$ from the same speaker as that of $u_i$, namely $a$.
Therefore, we propose a \emph{Speaker Highlighting Mechanism (SHM)}, with which we compute $|\mathcal{A}|$ different structured representations for each EDU such that each one highlights a specific speaker, where $\mathcal{A}$ is the set of all speakers in the dialogue.
This is particularly effective for multi-party dialogues.

Let $\bm{g}_{i,a}^S$ denote the structured representation of $u_i$ when highlighting speaker $a$, $p_i=u_j$ is the predicted parent of $u_i$, and $a_i$ is the speaker of EDU $u_i$. 
We compute the structured representations as follows:
\begin{equation}
\bm{g}_{i,a}^S=\left\{
\begin{array}{lcl}
\bm{0} & i=0\\
\mathbf{GRU}_{hl}(\bm{g}^S_{j,a}, \bm{h}_i \oplus \bm{r}_{ji} ) & a_i=a, i>0\\
\mathbf{GRU}_{gen}(\bm{g}^S_{j,a}, \bm{h}_i \oplus \bm{r}_{ji} ) & a_i\neq a, i>0\\
\end{array} \right.
\label{structured_representation}
\end{equation}
where $\oplus$ denotes vector concatenation, $\mathbf{GRU}$ stands for the functions of a GRU cell, and $\bm{r}_{ji}$ denotes the embedding vector of relation type $r_{ji}$, and $hl$ and $gen$ are short for $highlighted$ and $general$ respectively.

In Eq. (\ref{structured_representation}), $\bm{g}_{0,a}^S$ is set to a zero vector since the dummy root contains no real information. We compute $\bm{g}^S_{i,a}(i>0)$ based on the structured representation of its parent $\bm{g}^S_{j,a}$ that also highlights speaker $a$, and we use two different GRU cells $\mathbf{GRU}_{hl}$ and $\mathbf{GRU}_{gen}$ to respect whether the current speaker $a_i$ is highlighted or not. For the selected GRU cell, as shown in Figure \ref{structured}, $\bm{g}^S_{j,a}$ is the previous hidden state, $\bm{h}_i \oplus \bm{r}_{ji}$ is the input at the current step, and the new hidden state becomes $\bm{g}^S_{i,a}(i>0)$.

\subsection{Link Prediction and Relation Classification}

For each EDU $u_i$, the link predictor predicts its parent node $p_i$ and the relation classifier categorizes the corresponding relation type $r_{ji}$ if $p_i=u_j$. 
For each EDU $u_j(j<i)$ that precedes $u_i$ in the dialogue, we concatenate the representations $\bm{h}_i,\bm{g}^{NS}_i,\bm{g}^{NS}_j,\bm{g}^S_{j,a_i} $ to obtain an input vector $\bm{H}_{i,j}$ for link prediction and relation classification:
\begin{equation}
    \bm{H}_{i,j} = \bm{h}_i \oplus \bm{g}^{NS}_i \oplus \bm{g}^{NS}_j \oplus \bm{g}^S_{j,a_i}
    \label{H_input}
\end{equation}
For both $u_i$ and $u_j$, their non-structured global representations $\bm{g}^{NS}_i$ and $\bm{g}^{NS}_j$ are included in the input. 
We also add $\bm{g}^S_{j,a_i}$ which is the structured representation of $u_j$ when highlighting the speaker of $u_i$, namely $a_i$.
And since the structured representation of $u_i$ is unavailable at the current step, we add the local representation of $u_i$, namely $\bm{h}_i$ instead.

Taking $\bm{H}_{i,<i} \ (\bm{H}_{i,<i}=\bm{H}_{i,0},\cdots,\bm{H}_{i,i-1})$ as input, the link predictor estimates the probability that each $u_j(j<i)$ is the parent of $u_i$ on the dependency tree. The relation classifier then predicts the relation type between $u_j$ and $u_i$, if $u_j$ is the predicted parent of $u_i$.

\subsubsection{Link Prediction}

The link predictor first projects the input vectors $\bm{H}_{i,j}(j<i)$ to a hidden representation:
\begin{equation}
    \bm{L}_{i,j}^{link}=\tanh (\bm{W}_{link} \cdot \bm{H}_{i,j} + \bm{b}_{link} )
    \label{eq-link-prec}
\end{equation}
where $\bm{W}_{link}\in \mathbb{R}^{d_l\times d_h}$ and $\bm{b}_{link}\in \mathbb{R}^{d_h}$ are parameters, $d_l$ and $d_h$ are dimensions of $\bm{L}_{i,j}^{link}$ and $\bm{H}_{i,j}$ respectively.

The predictor then computes the probability that $u_j$ is the parent of $u_i$ on the predicted dependency tree as follows:
\begin{equation}
    o_{i,j}^{link}=\bm{U}_{link}\cdot \bm{L}_{i,j}^{link} + b_{link}' 
\end{equation}
\begin{equation}
    P(p_i=u_j|\bm{H}_{i,<i})=\frac{exp(o_{i,j}^{link})}{\sum_{k<i} exp(o_{i,k}^{link})}
    \label{link_prediction}
\end{equation}
where $\bm{U}_{link}\in\mathbb{R}^{1\times d_l}$ and $b_{link}'\in\mathbb{R}$ are also parameters.

Hence, the predicted $p_i$ is chosen as follows:
\begin{equation}
    p_i=\mathop{\rm argmax}_{u_j:j<i} P(p_i=u_j|\bm{H}_{i,<i}) 
\end{equation}

Unlike the local classifiers in \cite{li2014text,afantenos2015discourse,perret2016integer}, link prediction by $P(p_i=u_j|\bm{H}_{i,<i})$ depends on all candidate parents due to the $softmax$ normalization factor in Eq. (\ref{link_prediction}). 
During training, the gradient of each candidate parent is also dependent on all candidate parents, from which it can utilize more information for training, while other methods consider each candidate parent individually.

\subsubsection{Relation Classification}

Similar to link prediction, the relation classifier first projects the input vector $\bm{H}_{i,j}$ to a hidden representation, as follows:
\begin{equation}
    \bm{L}_{i,j}^{rel}=\tanh (\bm{W}_{rel} \cdot \bm{H}_{i,j} + \bm{b}_{rel} )
    \label{eq-rel-class}
\end{equation}  
where $\bm{W}_{rel}\in \mathbb{R}^{d_l\times d_h},\bm{b}_{rel}\in \mathbb{R}^{d_l}$ are parameters and $d_l$ equals to the  dimension of $\bm{L}_{i,j}^{rel}$.

The classifier then predicts the relation type $r_{ji}$ from the probability distribution over all types computed as follows:
\begin{equation}
    P(r|\bm{H}_{i,j}) = softmax(\bm{U}_{rel}\cdot \bm{L}_{i,j}^{rel} + \bm{b}_{rel}')
\end{equation} 
where $\bm{U}_{rel}\in\mathbb{R}^{K\times d_h},\bm{b}_{rel}'\in\mathbb{R}^K$ are also parameters.

\subsection{Loss Function}

We adopt the negative log-likelihood of the training data as the loss function:
\begin{equation}
    L_{link}(\Theta) = -\sum_{d\in \mathcal{D}} \sum_{i=1}^n \log P(p_i=p_i^{*}|\bm{H}_{i,<i})
\end{equation} 
\begin{equation}
    L_{rel}(\Theta) = -\sum_{d\in \mathcal{D}} \sum_{i=1}^n \log P(r_{ji}=r_{ji}^{*}|\bm{H}_{i,j},u_j=p_i^{*})
\end{equation}
\begin{equation}
    L_{all}(\Theta)=L_{link}(\Theta)+L_{rel}(\Theta)
\end{equation}
where $\Theta$ is the set of parameters to be optimized, $\mathcal{D}$ is the training data, $d$ is a dialogue in $\mathcal{D}$, $p_i^*$ and $r_{ji}^*$ are the golden parent and the corresponding golden relation type respectively.

Since the golden discourse structure is a dependency graph while our model predicts a dependency tree, to determine the golden parent $p_i^*$ of each EDU $u_i$ for training, we take the earliest EDU with a relation linking to $u_i$. And if $u_i$ is not linked from any preceding unit, we set $p_i^*=u_0$.

In $L_{rel}(\Theta)$, we use the log-likelihood of the relation type between $u_i$ and the golden parent $p_i^*$ rather than the predicted $p_i$, because the link predictor may predict incorrect $p_i$ such that the golden relation type between $p_i$ and $u_i$ can be unavailable.

\section{Experiments}

\subsection{Data Preparation}

We adopted the STAC Corpus \cite{STACcorpus}\footnote{\texttt{https://www.irit.fr/STAC/corpus.html}. We used the version released on March 21, 2018.} which is a multi-party dialogue corpus collected from an online game.
Its annotations follow Segmented Discourse Representation Theory (SDRT) \cite{asher2003logics}, where a discourse unit linked by a dependency relation may be an EDU, or a group of coherent discourse units named \emph{Complex Discourse Unit (CDU)} \cite{STACcorpus}.

\begin{figure}[ht]
    \centering
    \includegraphics[width=8cm]{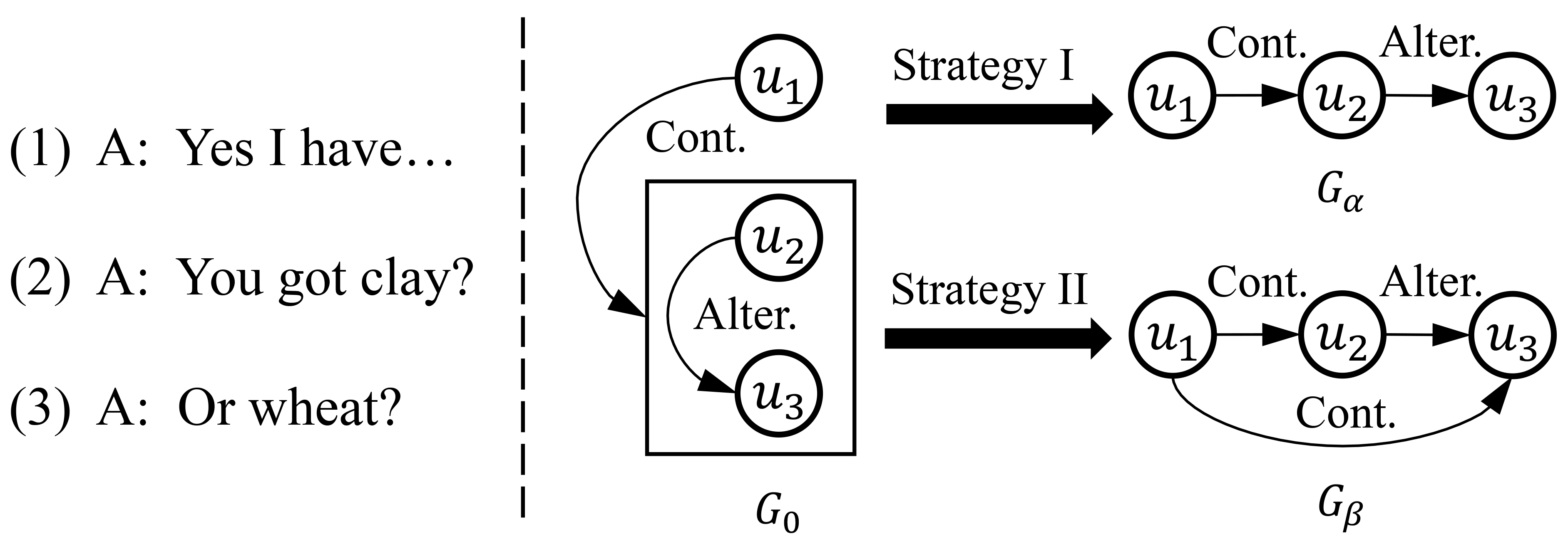}
    \caption{
    An example of eliminating a CDU which consists of $u_2$ and $u_3$ from the original SDRT structure $G_0$. ``Cont.'' is short for ``Continuation'' and ``Alter.'' for ``Alternation''. There are two strategies: \emph{Strategy I} links $u_1$ to the head of the CDU ($u_1\rightarrow u_2$), resulting in $G_{\alpha}$; and \emph{Strategy II} duplicates the link to every unit of the CDU ($u_1\rightarrow u_2;u_1\rightarrow u_3$), resulting in $G_{\beta}$.
    The relation $u_1\rightarrow u_3$ in $G_{\beta}$ is inappropriate, since $u_3$ is not a direct continuation of $u_1$.
    We took \emph{Strategy I} as it appears to be more reasonable in most cases.
    }
    \label{dataset}
\end{figure}

Previous studies for discourse dependency parsing have suggested that detecting CDUs remains challenging, and they thus transformed SDRT structures to dependency structures by eliminating CDUs \cite{muller2012constrained,afantenos2015discourse,perret2016integer}. 
Therefore, our task does not involve CDUs either. 
We adopted the strategy firstly proposed by \cite{muller2012constrained} to replace the CDUs with their heads recursively, where the head of a CDU is the earliest discourse unit in it without incoming relations. This strategy was also adopted by \cite{afantenos2015discourse,perret2016integer}.
But we did not apply another strategy mentioned by \cite{perret2016integer} which clones the relation to link every discourse unit in a CDU, since we found that this strategy brings many redundant and inappropriate relations as shown in Figure \ref{dataset}, and therefore, it may mislead the parsing models.

After eliminating the CDUs, the dataset consists of 1,062 dialogues, 11,711 EDUs and 11,350 relations in the training data; and 111 dialogues, 1,156 EDUs and 1,126 relations in the test data. We retained 10\% of the training dialogues for validation.
Similar to prior studies, for each dialogue, we manually added a relation from the dummy root to each EDU without an incoming relation, with a special relation type \emph{ROOT}.
Moreover, we discarded the dialogue act annotations on EDUs in the original dataset as they are irrelevant to our problem.

\subsection{Baselines}

We adopted the following baselines for comparison:

\begin{itemize}
    \item \textbf{MST} \cite{afantenos2015discourse}: A two-stage approach that adopts Maximum Spanning Trees (MST) to construct the discourse structure. MST builds a dependency tree using the probabilities from a dependency relation classifier which uses local information only.
    \item \textbf{ILP} \cite{perret2016integer} : A variant of MST which replaces the MST algorithm with Integer Linear Programming (ILP) to construct the discourse structure.
    \item \textbf{Deep+MST}: a variant of MST  which uses discourse representations from GRU encoders, instead of hand-crafted features or external parsers. 
    These representations are similar to those of our deep sequential model, but they only include non-structured representations.
    \item \textbf{Deep+ILP}: A variant of  ILP with the same modification as from MST to Deep+MST.
    \item \textbf{Deep+Greedy}: It is similar to Deep+MST and Deep+ILP, but this model adopts a greedy decoding algorithm which directly selects a parent for each EDU from previous EDUs with the largest probability.
\end{itemize}

We evaluated \emph{MST} and \emph{ILP} using the open source code from \cite{afantenos2015discourse,perret2016integer}\footnote{\texttt{https://github.com/irit-melodi/irit-stac}}. As for the deep baseline models, since the structured representations are unavailable, we replaced $\bm{g}^S_{j,a_i}$ in $\bm{H}_{i,j}$ with $\bm{h}_j$ instead, and thus the input for dependency prediction becomes:
\begin{equation}
    \bm{H}_{i,j}^{'} = \bm{h}_i \oplus \bm{g}^{NS}_i \oplus \bm{h}_j \oplus \bm{g}^{NS}_j
    \label{H_input_baseline}
\end{equation}
where we concatenate the non-structured representations of EDU $u_i$ and $u_j$ together.
Moreover, to compare the models fairly, the dimensions of the input vector in all the deep baseline models and our sequential model are kept the same.

\subsection{Implementation Details}

The word vectors are initialized with 100-dimensional Glove vectors \cite{pennington2014glove} and are fine-tuned during training. 
The dimensions of the relation embeddings an the  discourse representations are set to 100 and 256 respectively.
And the dimensions of the hidden representations in link prediction and relation classification are set to 512.
Dropout is adopted before the input of each GRU cell, with a probability of 0.5.
We use Stochastic Gradient Descent (SGD) to train the model, with the mini-batch size set to 4. The initial learning rate is set to 0.1 and it decays at a constant rate of 0.98 after each epoch.

Moreover, we experimented with two settings of our deep sequential model. One is a shared version where the link predictor and the relation classifier share the same input vector $\bm{H}_{i,j}$. The other is a non-shared version where the two input vector $\bm{H}_{i,j}$ in Eq. (\ref{eq-link-prec}) and that in Eq. (\ref{eq-rel-class}) are from networks with different parameters respectively. We finally took the later one and also applied it to deep baseline models.

\subsection{Results}

\begin{table}[ht]
  \centering
  \begin{tabular}{l|c|c}
    \hline
    Model & Link & Link \& Rel\\
    \hline
    MST & 68.8 & 50.4\\ 
    ILP & 68.6 & 52.1\\
    Deep+MST & 69.6 & 52.1 \\
    Deep+ILP & 69.0 & 53.1 \\
    Deep+Greedy & 69.3 & 51.9 \\
    \hline
    Deep Sequential (shared) & 72.1 & 54.7 \\
    \textbf{Deep Sequential} & \textbf{73.2} & \textbf{55.7} \\
    \hline
  \end{tabular}
  \caption{$F_1$ scores (\%) for different models. {\it Link} means link prediction; and {\it Link \& Rel} means that a correct prediction must predict dependency link and relation type correctly at the same time.}
  \label{main_results}
\end{table}

We adopted micro-averaged $F_1$ score as the evaluation metric.
Results for different models are shown in Table \ref{main_results}, where ``Link'' denotes link prediction while ``Link \& Rel'' stands for that a correct prediction must predict dependency link and relation type correctly at the same time.

Our deep sequential model outperforms all the baselines significantly (bootstrap test, $p<0.05$), demonstrating the benefit of predicting dependency relations and constructing the discourse structure jointly and alternately.
Besides, we observed that the performance drop when link prediction and relation classification share the same discourse representations (\emph{Deep Sequential (Shared)}).
This is probably because that it is hard to train the discourse encoders to simultaneously capture the information needed by both link prediction and relation classification.

Moreover, compared to \emph{MST} and \emph{ILP} that rely on hand-crafted features and external parsers, the deep baseline models \emph{Deep+MST} and \emph{Deep+ILP} achieve higher $F_1$ scores. This demonstrates that discourse representations from hierarchical GRU encoders are more effective than traditional features.
\emph{Deep+Greedy} has a lower $F_1$ score on \emph{Link \& Rel} compared to \emph{Deep+MST} and \emph{Deep+ILP}, indicating that more sophisticated decoding algorithms help construct better structures. Interestingly, our deep sequential model, which does not have any complex decoding algorithm, still outperforms those baselines. This further validates the effectiveness of our sequential model.

\begin{figure*}[ht]
    \centering
    \includegraphics[width=16cm]{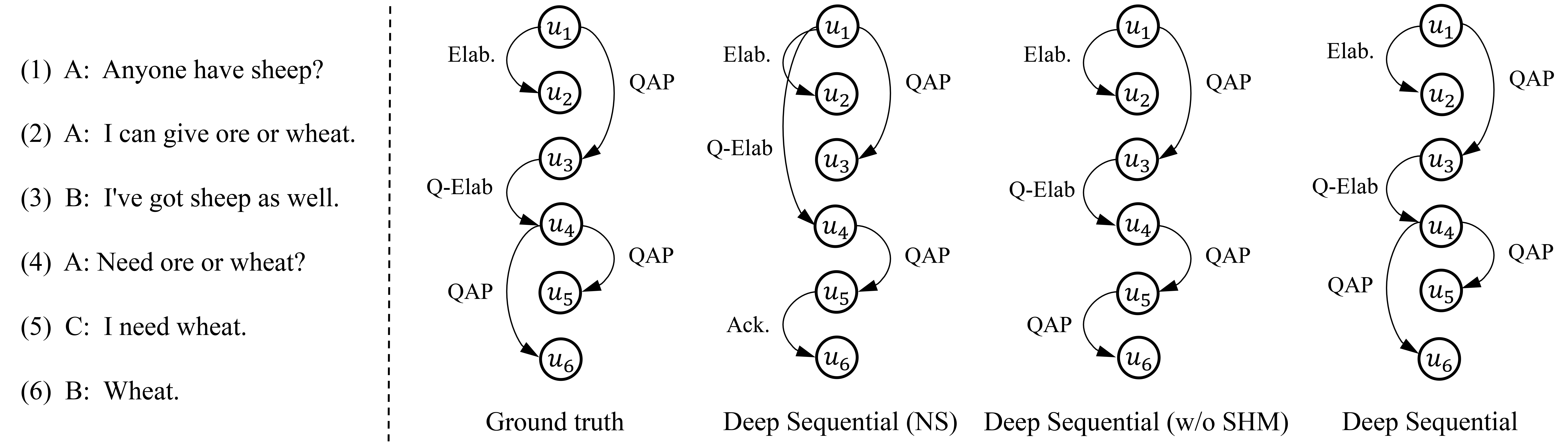}  
    \caption{A dialogue example from three speakers, along with the golden discourse structure and discourse structures predicted by various models. ``Elab.'' is short for ``Elaboration'', ``QAP'' for ``Question-Answer-Pair'', ``Q-Elab'' for ``Question-Elaboration'', and ``Ack.'' for ``Acknowledgement''. $u_i$ in the graphs corresponds to the $i$-th utterance in the left panel.}
    \label{case}
\end{figure*}

\subsection{Effectiveness of the Structured Representations}

To evaluate the effectiveness of the structured representations, we devised the following three variants of our deep sequential model for comparison:

\begin{itemize}
    \item \textbf{Deep Sequential (NS)}:
    We removed the structured representations from our original deep sequential model.
    Similar to that of other deep baselines, the input to the link predictor and relation type classifier has only non-structured representations, as defined by Eq. (\ref{H_input_baseline}).
    
    \item \textbf{Deep Sequential (Random)}:
    In this variant, both non-structured and structured representations are used, but the structured representations encode a random structure. For each EDU, we randomly choose a parent from the preceding EDUs and a random relation type, to obtain its structured representation.
    
    \item \textbf{Deep Sequential (w/o SHM)}:
    We disabled the speaker highlighting mechanism in the full model to evaluate the effectiveness of this mechanism.
\end{itemize}

\begin{table}[ht]
  \centering
  \begin{tabular}{l|c|c}
    \hline
    Model & Link & Link \& Rel \\
    \hline
    Deep+Greedy & 69.3 & 51.9 \\
    \hline
    Deep Sequential (NS) & 71.0 & 53.7 \\
    Deep Sequential (Random) & 71.8 & 53.7 \\
    Deep Sequential (w/o SHM) & 71.7 & 54.5 \\
    \hline
    \textbf{Deep Sequential} & \textbf{73.2} & \textbf{55.7} \\
    \hline
  \end{tabular}
  \caption{$F_1$ scores (\%) for different models. }
  \label{results_variants}
\end{table}

Results in Table \ref{results_variants} reveal the following observations:

\begin{enumerate}

\item Our full model (\textit{Deep Sequential}) outperforms \emph{Deep Sequential (NS)} and \emph{Deep Sequential (Random)}, indicating that the structured representations which encode the predicted discourse structure are crucial for dependency relation prediction.
    
\item When the speaker highlighting mechanism is disabled (\emph{Deep Sequential (w/o SHM)}), there is a remarkable drop on performance, which demonstrates that the speaker highlighting mechanism can improve the prediction of dependency relations.

\item A random structure can help link prediction slightly, as can be seen from the comparative results between \emph{Deep Sequential (Random)} and \emph{Deep Sequential (NS)} ($71.8$ vs. $71.0$).
However, \emph{Deep Sequential} is much better than \emph{Deep Sequential (Random)}, indicating that structured representations can effectively encode valuable information from the predicted discourse structure.
\end{enumerate}    
    
We also noticed that \emph{Deep Sequential (NS)} still has an improvement over \emph{Deep+Greedy}, even without structured representations.
It is because the probabilities of dependency links are dependent on all candidate parents in the sequential models due to the $softmax$ normalization, whereas the baseline models predict each dependency link individually.
Thereby, the global information from other candidate links benefits the training of sequential models.

\subsection{Case Study}
We provide an example to show how structured information helps the model to better understand the development of the dialogue, which is important for dependency prediction.

As shown in Figure \ref{case}, \emph{Deep Sequential (NS)} incorrectly predicts the parent of $u_4$ as $u_1$ while the ground truth is $u_3$, yet both \emph{Deep Sequential (w/o SHM)} and \emph{Deep Sequential} make correct predictions.
The previously predicted dependency relation $u_1\rightarrow u_3$ with type \emph{QAP}, is encoded by structured representations.
It thus helps the model to understand that the question in $u_1$ has been answered by $u_3$, and therefore, it is more likely that $u_4$ responds to $u_3$, rather than elaborates the original question $u_1$ which has already been responded by others.

Moreover, both \emph{Deep Sequential (NS)} and \emph{Deep Sequential (w/o SHM)} incorrectly predict the parent of $u_6$ while \emph{Deep Sequential} makes a correct prediction. 
Thanks to the speaker highlighting mechanism, when predicting the parent of $u_6$, the model highlights the previous EDU $u_3$ from the same speaker as that of $u_6$ (i.e. speaker $B$) in the structured representations.
Thereby, in order to predict a dependency relation $u_4\rightarrow u_6$, the model tends to utilize the information that 
it is $u_4$ which responds to the previous EDU $u_3$ by the speaker of $u_6$.

\section{Conclusion and Future Work}

In this paper, we propose a deep sequential model for discourse parsing on multi-party dialogues. The model predicts dependency relations and builds the discourse structure jointly and alternately. It decides the dependency links between EDUs and the corresponding relation types sequentially, utilizing the structured representation of each EDU encoded with a structured encoder, and in return, the predicted dependency relations are used to construct the discourse structure incrementally.
Experiments show that our sequential model outperforms all the state-of-the-art baselines significantly and the structured representations can effectively improve dependency prediction.

We not only propose an approach to parse discourse structures, but also an approach to utilize them via a structured encoder. We have further demonstrated the benefit of discourse structures. As future work, our method can be enhanced and applied to improve approaches for other NLP tasks of multi-party dialogues.

\section{Acknowledgments}

This work was jointly supported by the National Science Foundation of China  (Grant No.61876096/61332007), and
the National Key R\&D Program of China (Grant No. 2018YFC0830200). We would like to thank Prof. Xiaoyan Zhu for her generous support.

\bibliography{paper}
\bibliographystyle{aaai} 

\end{document}